\documentclass{article}
\usepackage{ijcai21}

\usepackage{times}
\usepackage{soul}
\usepackage{url}
\usepackage[hidelinks]{hyperref}
\usepackage[utf8]{inputenc}
\usepackage[small]{caption}
\usepackage{graphicx}
\usepackage{amsmath}
\usepackage{amsthm}
\usepackage{caption}
\usepackage{booktabs}
\usepackage{algorithm}
\usepackage{algorithmic}
\usepackage{lipsum}
\usepackage{mathtools}
\usepackage{tabularx}
\usepackage{cuted}
\usepackage{amssymb}
\usepackage{varwidth}
\usepackage{subcaption}
\title{HydroDeep – A Knowledge Guided Deep Neural Network for Geo-Spatiotemporal Data Analysis}

 \author{
    Aishwarya Sarkar$^1$
    \and
    Jien Zhang$^2$\and
    Chaoqun Lu$^{3}$\and
    Ali Jannesari$^4$
    \affiliations
    $^{1,4}$Department of Computer Science, Iowa State University\\
    $^{2,3}$Department of Ecology, Evolution, and Organismal Biology, Iowa State University
    \emails
    \{asarkar1, jienz, clu, jannesar\}@iastate.edu
 }

\begin{document}
\maketitle
\begin{abstract}
    Due to limited evidence and complex causes of regional climate change, the confidence in predicting fluvial floods remains low. Understanding the fundamental mechanisms intrinsic to geo-spatiotemporal information is crucial to improve the prediction accuracy. This paper demonstrates a hybrid neural network architecture - HydroDeep, that couples a process-based hydro-ecological model with a combination of Deep Convolutional Neural Network (CNN) and Long Short-Term Memory (LSTM) Network. HydroDeep outperforms the independent CNN's and LSTM's performance by 1.6\% and 10.5\% respectively in Nash–Sutcliffe efficiency. Also, we show that HydroDeep pre-trained in one region is adept at passing on its knowledge to distant places via unique transfer learning approaches that minimize HydroDeep's training duration for a new region by learning its regional geo-spatiotemporal features in a reduced number of iterations.
\end{abstract}
\section{Introduction}
Deep Learning (DL) has found immense success in many commercial applications such as natural language processing, and computer vision due to their ability in learning complex dependencies from the data. However, many underlying factors lead to one consequence in the real world. With inadequate information about an environment's underlying physical mechanisms, highly data-driven, DL is inclined to false predictions since it is challenging to capture highly complex relationships entirely from data. Recent work shows an increase in DL applications in hydrology because DL is the only bet at processing dependencies in extensive hydrological training data \cite{sit2020comprehensive}. Nevertheless, due to DL algorithms' limited knowledge of underlying process-based (PB) mechanisms, PB hydrological models are necessary to study environmental systems. However, most PB models although based on fundamental laws of a system, are often mere approximations of reality due to omitting processes to maintain computational efficiency. The predictions solely rely on the qualitative parameterization of environment drivers - climate, soil property, and land cover.

This paper focuses on grid-based modeling of physical systems. In particular, we aim to build a network that can capture regional geo-spatiotemporal features influencing a catastrophic event of a flood. Flood events are the most frequent type of natural disasters worldwide. The recent surge in population growth and rapid urbanization have made it extremely essential to tackle this issue. Every year, farmlands and urban infrastructures go through significant and sometimes irreparable damages due to flash floods. In 2019, about a quarter-million acres of fertile farmland was underwater for four months in the Mississippi Delta region \cite{nytimes2019:scheme}.

Although in hydrological sciences, modeling of rainfall-runoff using approaches of regression date back 170 years \cite{beven2011rainfall,mulvaney1851use}, recent researches have incorporated more profound concepts of physical properties of the catchments, boundary conditions, and spatial variability into mathematical model formulations. Nonetheless, a large degree of uncertainty with the current approaches remains, making the quantitative estimation of climate change impacting extreme rainfall events an active research area. 

In this paper, we propose to couple a PB hydro-ecological model with a hybrid neural network architecture - HydroDeep, to analyze grid-based geo-spatiotemporal features of a watershed contributing to local river discharge. In our experiments, HydroDeep outperformed the Nash–Sutcliffe efficiency of Convolutional Neural Networks (CNN) by 1.6\% and Long  Short-Term  Memory Networks (LSTM) by 10.5\%. We also propose a new application area of transfer learning to analyze similarities and dissimilarities in geo-spatiotemporal characteristics of watersheds while reducing the extensive training time required in training local DL models. In summary, our contributions are:
\begin{itemize}
\item We introduce HydroDeep - a novel hydrological knowledge powered deep neural network that processes grid-based spatiotemporal inputs to predict regional river discharge measurements.
\item For more accurate prediction, we introduce an approach to utilize the prediction day's input measurements.
\item We propose a new transfer learning application in analyzing geo-spatiotemporal features of different regions while reducing extensive training time and large-scale dataset availability, required otherwise.
\end{itemize}

\section{Background and Related Works}PB hydrological models have a history dating back to the 1960s \cite{crawford1966digital} and are often criticized as overly complex and difficult to use. On the contrary, researchers also believe that these models have a massive advantage in situations where knowledge of distributed state variables and physical constraints is essential. In recent times, many hydrological models are still predominantly used in the real world; it is irreplaceable at capturing any environment's PB mechanisms. Simple hydrological models have proved their worth at being useful and elegant in describing large-scale patterns. For example, applying fundamental physical principles such as Maximum Entropy Production or Maximum Energy Dissipation has been used to explain the Earth system, and hydrological processes \cite{kleidon2009thermodynamics,wang2009model}. However, understanding a system's general organization does not solely provide insights into how the principal variables interact over space and time. \citeauthor{fatichi2016overview} discuss all the prevalent challenges of distributed process-based models in hydrology.

Preliminary researches in the applications of deep learning techniques in hydrology date back to the 1990s, where Artificial Neural Networks (ANNs) were first used for rainfall-runoff prediction \cite{daniell1991}. Due to ANN's limitation in retaining the temporal inputs' sequential order, Recurrent Neural Networks (RNNs) were introduced. Although RNNs have proved to be more distinguished at using sequential information in time-series data but still had inhibition in remembering long-term dependencies. Hence, Long Short-Term Memory Networks (LSTMs) were proposed to learn long-term temporal dependencies, and they outperformed primitive ANNs in rainfall-runoff simulations \cite{Hu_2018}. Several recent studies have started coupling PB models with DL networks and claimed that hydrological model's knowledge has been useful in real-time forecasts \cite{yang2019real}. \citeauthor{sit2019decentralized} used Gated Recurrent Units (GRU) in their study and claimed that LSTM-seq2seq with GRUs outperformed the individual state of the art performance in the Clear Creek Watershed studies in Iowa \cite{xiang2020distributed}. Along with LSTMs, 1D Convolutional Neural Networks (CNNs) also showed promising performance in rainfall-runoff modeling where two parallel convolution filters processed separate time series allowing faster processing of data \cite{van2020deep}. 
A combination of CNN and LSTM, named initially as Long-term Recurrent Convolutional Network (LRCN), has been recently developed and shows promising results in generating textual descriptions of images \cite{donahue2015long}. With computer vision, the application of CNN-LSTM also spans over other application areas such as speech recognition and natural language processing \cite{vinyals2015show}. Results from a CNN-LSTM network (APNet) outperformed all the traditional deep learning models commonly used in time-series analysis \cite{Huang_2018}.

\section{Approach}The ability of 2D and 3D CNNs, widely used to recognize spatial patterns in image classification problems, remains intact in its one-dimensional variation (1D-CNN). Being highly noise-resistant models, 1D-CNNs can extract very informative and in-depth features independent of time and are more often used in natural language processing. When applied to the 1D time-series data, the convolutional layers essentially map the internal features of a temporal sequence. The motive behind the integration of CNN and LSTM in HydroDeep lies in capturing both the spatial and temporal dependencies of a watershed. The convolutional layers help extract local geospatial features between the input variables and pass them to LSTMs to support temporal sequence prediction.

\subsection{The Process-based Model - DLEM}
The Dynamic Land Ecosystem Model (DLEM) that we use in this work is a  PB hydro-ecological model that mimics the plant physiological, biogeochemical, and hydrological processes in the plant-soil-water-river continuum \cite{liu-et-al:scheme,lu2013net:kl-one}.
\begin{figure}[b]
\centering
\includegraphics[width=0.9\columnwidth]{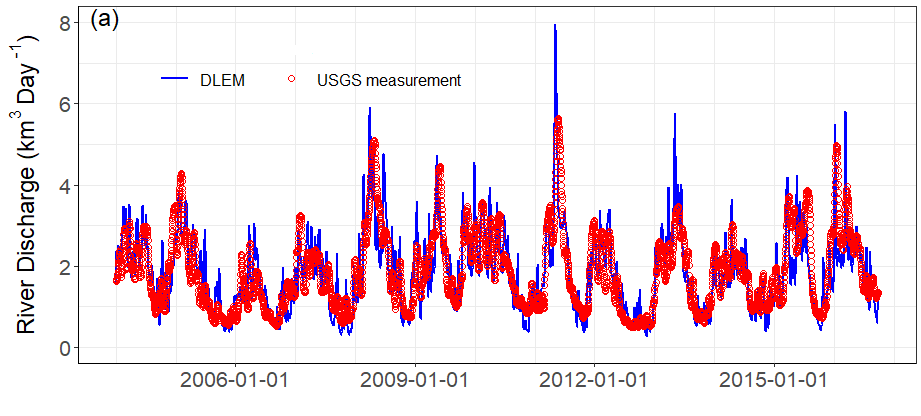} 
\caption{Comparison of DLEM-modeled and USGS-monitored daily river discharge from the Mississippi and Atchafalaya river basin (MARB). The daily river discharge is from April 1, 2004, to September 28, 2016. The USGS river discharge was calculated based on the daily river discharge measurements at the Melville gauge station (station ID 07381495), measuring the discharge of Atchafalaya River and St. Francisville gauge station (station ID 07373420) measuring the discharge of the Mississippi River.}
\label{fig1}
\end{figure}
\begin{figure*}[ht]
\centering
\includegraphics[width=0.9\textwidth]{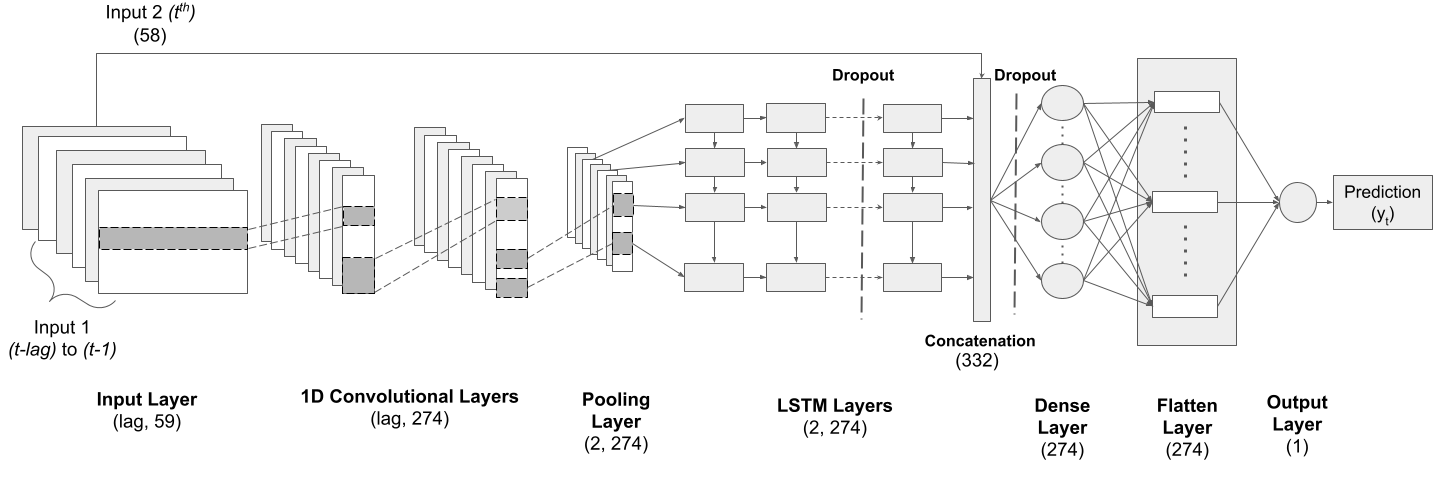} 
\caption{HydroDeep Network Architecture}
\vspace{-2mm}
\label{fig2}
\end{figure*}
In DLEM, river network and in-stream processes are incorporated into a terrestrial modeling scheme. The design of grid-to-grid connection tracks significant features of a region, including within-grid heterogeneity, grid-to-grid flow, and land-aquatic linkage. DLEM models the water movement from land to aquatic systems at a daily time step with each grid cell comprising multiple land cover types, rivers, and lakes with their area percentage prescribed by land-use history data \cite{Lu2020}. Research shows that DLEM has been extensively validated against measurements from LTER, NEON, Ameri Flux, USDA crop yield survey, and USGS gauge monitoring and are widely used to quantify the spatiotemporal variations in the pool and fluxes of water, carbon, and nitrogen coupling (water-C-N) at the site, and regional scales \cite{liu-et-al:scheme,lu2013net:kl-one,yang2015increased,yu2019largely,lu2019-et-al:scheme}. The preliminary results from DLEM at the outlet of the Mississippi and Atchafalaya river basin (MARB) show that the variations of daily river discharge are very close to the USGS observed river discharge over the years (Figure 1) \cite{Lu2020}. However, the DLEM capability in capturing very high daily river discharge is still limited (Figure 1).
\subsection{HydroDeep}
\subsubsection{Problem Statement}If ~$g_i=\{g_1, g_2, \dots, g_L\}$ is the spatial grid vector where L is the total number of grids covering a region having spatial coordinates ~$c_i=\{x_{g_1},y_{g_1}\}, \{x_{g_2},y_{g_2}\}, \dots, \{x_{g_L},y_{g_L}\}$, the distance to these grids from the nearest river or water source is ~$d_i=\{d_{g_1}, d_{g_2}, \dots, d_{g_L}\}$. On a certain day $t$, the grids have precipitation measurements $p_{g_i}$ where $p_{g_i}=\{p_{g_1}, p_{g_2}, \dots , p_{g_L}\}$ which are mapped to their corresponding grid-based DLEM simulated runoff, $r_{g_i}$ where $r_{g_i}=\{r_{g_1}, r_{g_1}, \dots , r_{g_L}\}$. The extent to which each grid's precipitation contributes to the river discharge  depends  extensively  on  the  grid’s  distance  to  the nearest water source. The distance vector is thus transformed to a distance weight vector $\widetilde{d}_{g_i}$ such that higher distance weights are applied to $p_{g_i}$ if the said grid is closer to a local river thus contributing more to the regional river discharge (Supplementary Section 1.).  We denote weighted precipitation vector as $\widetilde{p}_{g_i}=\widetilde{d}_i \odot {p}_{g_i}$ . At time t, the input vectors $\widetilde{p}_{g_i, t}$ and $r_{g_i, t}$ are labelled with daily river discharge observations ~$D_t = \{D_1, D_2, \dots, D_n\}$ where $n$ is the total number of river discharge observations. Eq. 1 shows input vector $x_t$. 
\begin{equation}
    x_t = f[(\widetilde{p}_{g_i, t}), (r_{g_i, t}) ]\\
\end{equation}
From a multivariate time-series point of view, we can define the problem more formally. Given an input sequence of time-series signals $X=(x_1, x_2, \dots, x_t)$ with $x_i\in\mathbb{R}^n$, where the variable dimension is $n$, we want to predict the corresponding target outputs $Y=(y_1, y_2, \dots ,y_t)$. The aim is to obtain a non-linear mapping between $X$ and $Y$.
\subsubsection{Input Data}
The precipitation measurement of the target prediction day $(t)$ may trigger an event of high river discharge measurement on the very same day. A continuous drought for days can be followed by a hurricane, leading to a flood overnight, making it crucial to include $t$'s input drivers into the network. Therefore, HydroDeep takes inputs at two different network stages to accommodate day $t$'s inputs. HydroDeep also uses the ``lag" parameter exclusively to look back at a period of days and extract temporal features from the historical data. The first input contains measurements from $(t-lag)$ to $(t-1)$ days. The second input has the measurements of $t^{th}$ day and gets concatenated halfway through the network.
\subsubsection{Workflow}
Since watersheds vary in area, each watershed has a specific number of grids $(L)$. w13 has 29 grids with two input variables located at each grid. Daily river discharge observations are given as labels $(y)$ to all 29 grids. As previously mentioned, the network takes a $lag$ value, and hence the Input 1's shape with 2L inputs and 1 label per time step becomes $(lag, 2L + 1)$. HydroDeep has an initial input layer that is customizable to different input shapes to make our model adaptive to other watersheds. Figure 2 shows the overall framework of HydroDeep. At first, convolutional layers are used on Input 1 to extract local (spatial) temporal features. A max-pooling layer follows these layers to reduce the dimensions of the feature maps by only selecting the salient features. The obtained vectors are then fed into LSTM's input gate units. Each LSTM layer has three gate units - input, output, and forget gate that determine the state of each memory cell through multiplication operations. The memory units preserve long-term memory by updating their states as each gate unit gets activated. The hidden state of the LSTM cells is $h_t$ at every time-step $t$. We have four stacked LSTM layers with a subsequent Dropout that regularizes the network.  
\begin{align}
    f_t=\sigma(W_f\cdot[h_{t-1},x_t]+b_f)\\
    i_t=\sigma(W_i\cdot[h_{t-1},x_t]+b_i)\\
    o_t=\sigma(W_o\cdot[h_{t-1},x_t]+b_o)
\end{align}
Equations 2, 3, and 4 show the mathematical operations of the forget gate, the input gate, and the output gate, respectively where $f$, $i$ and $o$ are the output of the operations at time-step t. $\sigma$ is the non-linear $sigmoid$ activation function used at the gates and only allows values between $[0, 1]$. The weight matrix of each unit gate is $W$, and their bias vector is $b$. The cell states and the hidden states are formulated from the forget, the input, and the output gate and are shown in Equations 5 and 6, respectively. The cell state, defined as $C_t$ uses $tanh$ to restrict its values between -1 and 1. This is multiplied by the output of the sigmoid gate $o_t$ to update the hidden state.
\begin{align}
    C_t=f_t * C_{t-1} + i_t*\tanh(W_C\cdot[h_{t-1},x_t]+b_C)\\
    h_t=o_t * \tanh(C_t)
\end{align}
At this stage, the network gets Input 2, which is concatenated with the LSTM layer's output $(h_t)$, and are processed collectively by a fully connected layer. The final layer is the output layer, which predicts the river discharge measurement at $t$. In the training phase, we use Adam Optimization Algorithm to backpropagate the model parameters.

\subsection{Transfer Learning}An event of flood or drought is heavily dependent on environmental drivers. Likewise, each of these drivers' inherent local spatiotemporal patterns is bound to be unique based on their geographical location. Consequently, one model that has learned the local spatiotemporal patterns of a region will fail to perform accurately for a geographically distant region with different characteristics.As a result, the model should be retrained perpetually for every new unique region.  Alternatively, a global model can be trained on a larger area covering many watersheds to address this problem, but it will fail to capture the local patterns. Besides, both these methods are expensive as they require more training time and computational power to train such a model of global extent. Therefore, we use Transfer Learning to reuse HydroDeep's knowledge from one region to another. More formally, transfer learning consists of a domain $D$ and a task $T$ where the domain $D$ is the marginal probability distribution $P(X)$ over a feature space $X = \{x_1, x_2,\dots, x_n\}$. Given a domain $D = \{X, P(X)\}$, a task $T$ consists of a conditional probability distribution $P(Y|X)$ over a label space $Y$. The conditional probability distribution is usually learned from the pairs $\{x_i, y_i\}$ in the training samples where $x_i \in X$ and $y_i \in Y$. Suppose there is a source domain $D_s$ with a source task $T_s$ and a target domain $D_t$ with a target task $T_t$, through transfer learning we try to learn the target conditional probability distribution $P(Y_t| X_t)$ in target domain $D_t$, from the knowledge learned from $D_s$ and $T_t$.\cite{laptev2018applied}

For our experiment, we chose the Thompson Fork Grand River basin at David City (watershed 13) as our source domain and transferred its knowledge to 5 target domains - West Nodaway River  near Shambaugh (watershed 14), East Nishnabotna River near Shenandoah (watershed 15), Turkey River near Garber (watershed 4), South Skunk River near Oskaloosa (watershed 10), and Rock River near Hawarden (watershed 23) \cite{jones-et-al:scheme}. We evaluated four transfer learning approaches and we present the detailed findings in Section 5.4.

\section{Experimental Design}\subsection{Dataset} We pulled the dataset from the U.S. Geological Survey's (USGS) daily discharge measurements for Iowa Streams having daily discharge measurements of 23 watersheds covering the state of Iowa (Figure 3) \cite{jones-et-al:scheme}. The red triangles in the figure are where the USGS recorded their measurements. For our initial study, we chose watershed 13 (w13) due to its smaller size.  Alongside, we use daily precipitation at a 5-arc-min resolution generated from high-resolution gridded meteorological data products from station observations by the Climatic Research Unit (CRU) of the University of East Anglia \cite{doi:10.1002/joc.1181}, and North America Regional Reanalysis (NARR) dataset from a combination of modeled and observed data \cite{10.1175/BAMS-87-3-343}. We also use DLEM-simulated surface and subsurface runoff to guide our DL network \cite{Lu2020}. Climate, land management, and environmental drivers steer DLEM simulation and are used to represent our ``best estimate" of land-to-aquatic surface and subsurface runoff across the watershed.
\begin{figure}[t]
\centering
\includegraphics[width=0.9\columnwidth]{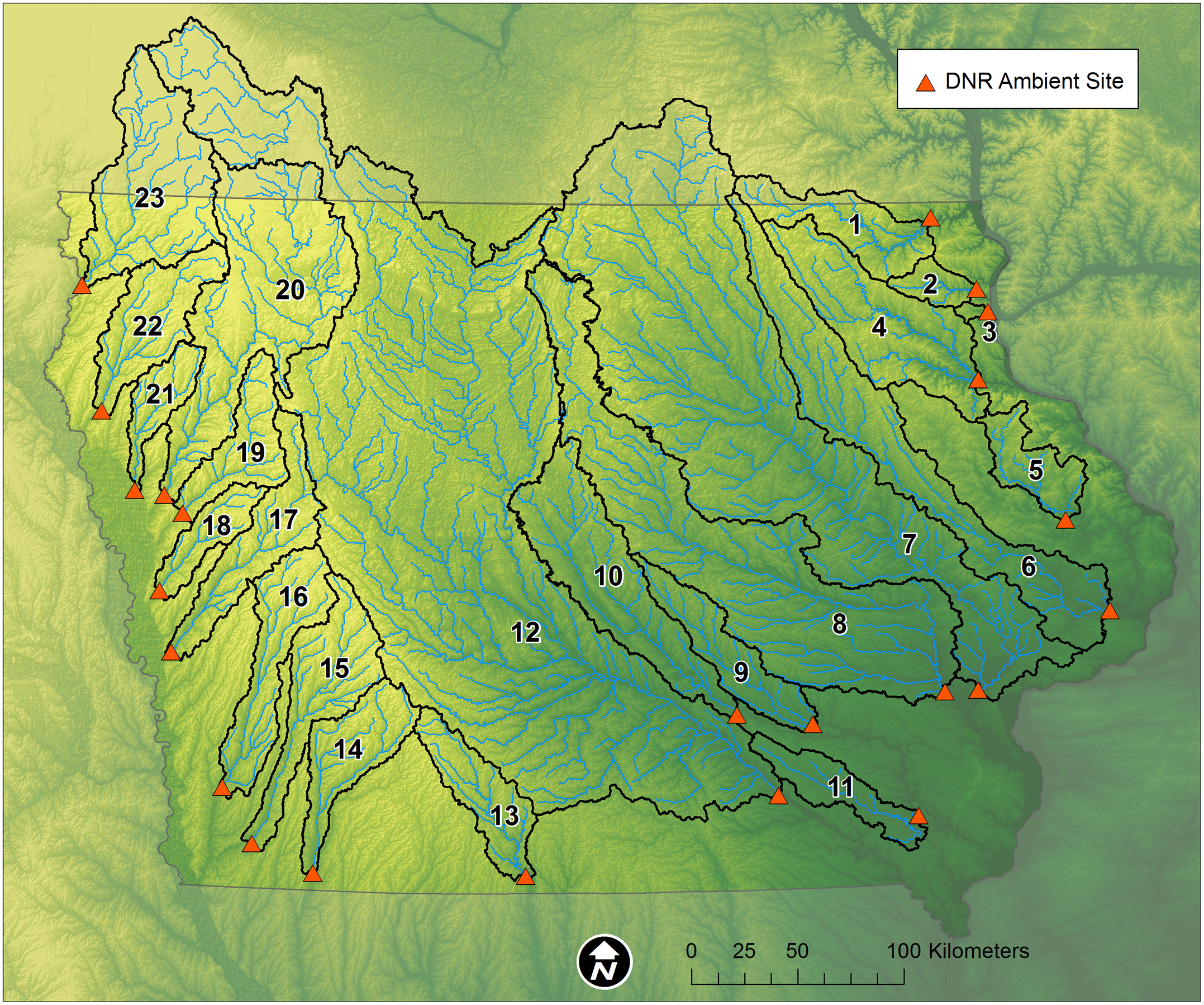} 
\caption{Iowa Stream Sites and observation regions. The daily river discharge observations are recorded at each red triangle. The numbers correspond to watersheds.}
\vspace{-3mm}
\label{fig3}
\end{figure}
\begin{figure*}[t]
\centering
\includegraphics[width=\textwidth]{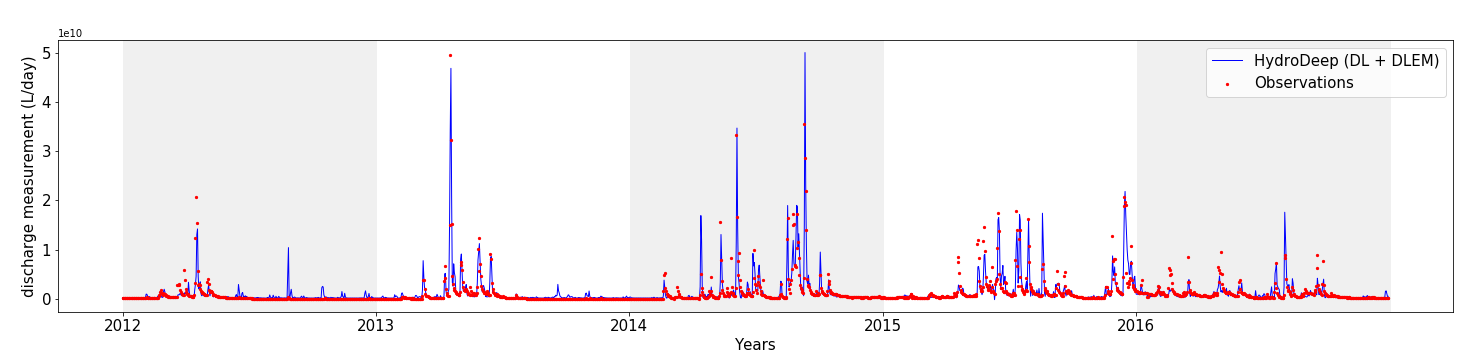} 
\caption{The HydroDeep predicted daily river discharge of w13 is plotted against river gauge measurements from Jan 1, 2012 to Dec 31, 2016.}
\vspace{-1mm}
\label{fig4}
\end{figure*}
\subsection{Model Parameter Settings}The distance weighted input vectors are first normalized and transformed into time-series samples using lag or lookback parameter. We split the dataset into training set and test set (7:3), using 25\% of the training samples for validation. The training set contains measurements and observations from Jan 1, 2000 to Dec 31, 2011. We keep data from the period of Jan 1, 2012 to Dec 31, 2016 for evaluation. The network uses the $tanh$ activation function on the convolutional and recurrent layers and $relu$ on the fully connected layer. Adam optimizer is used to further tune the model parameters.
\section{Results and Discussion}\subsection{Evaluation Metrics}We selected three model evaluation techniques of watershed simulations from \citeauthor{moriasi2007model}, Nash–Sutcliffe efficiency (NSE), RMSE-observations standard deviation ratio (RSR), and Percent bias (PBIAS). The respective equations are shown in Equation 7, 8, and 9 where $y_{i}^{obs}$ is the $i^{th}$ river discharge observation, $y_{i}^{sim}$ is the corresponding Hydrodeep simulated measurement, $y^{mean}$ is the mean of the observation values and $n$ is the total number of observations.
\begingroup
\allowdisplaybreaks
\begin{align}
    NSE = \left[1 - \frac{\sum\limits_{i=1}^{n}{(y_{i}^{obs} - y_{i}^{sim})^2}}{\sum\limits_{i=1}^{n}{(y_{i}^{obs} - y^{mean})^2}}\right]\\
    PBIAS = \left[ \frac{\sum\limits_{i=1}^{n}{(y_{i}^{obs} - y_{i}^{sim})^2} * 100}
    {\sum\limits_{i=1}^{n}(y_{i}^{obs})}\right]\\
    RSR = \frac{RMSE}{STDEV_{obs}} = \left[\frac{ \sqrt{\frac{1}{n}\sum\limits_{i=1}^{n} {(y_{i} ^ {obs} - y_{i} ^ {sim})^2}}} {\sqrt{\frac{1}{n}\sum\limits_{i=1}^{n}{(y_{i} ^ {obs} - y ^ {mean})^2}}}\right]
\end{align}
\endgroup
\subsection{Hyperparameter Tuning} The hyperparameters of HydroDeep have been tuned to achieve the best performance (Supplementary Section 2.). Out of all the hyperparameters, HydroDeep was most susceptible to the lookback $(lag)$ parameter. In the first iteration, the lookback window sets itself from the start day of the training sample-set and then slides over with a step value of 1 day per iteration for the following iterations. We experimented with lookback values varying from 3 to 11 days and observed that increasing the number of days did not consistently improve the performance. The value was set manually before each run to observe the NSE performance (Figure 5). From our analysis, a lookback period of 7 days in HydroDeep has achieved the best performance in NSE throughout.
\begin{figure}[h]
\centering
\includegraphics[width=0.9\columnwidth]{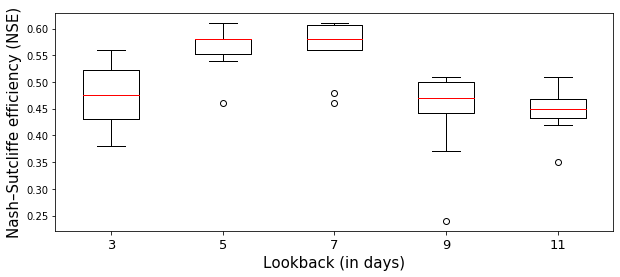} 
\caption{NSE of different model runs on watershed 13 with various lookback $(lag)$ values in HydroDeep.}
\vspace{-3mm}
\label{fig5}
\end{figure}
\subsection{Prediction Performance} After the training is complete, Hydrodeep's performance is evaluated on the test dataset (Figure 4). Figure 6 shows an arbitrarily chosen window of 3 months from 2013 in w13. We compare the prediction power of the three approaches, DL which represents a similar neural network architecture as HydroDeep but without DLEM simulated runoff as input, standalone DLEM, and our proposed HydroDeep, against the observations to evaluate where they stand in comparison. According to our results, Hydrodeep performs better than its counterparts. 

HydroDeep on w13 achieves an  NSE of 0.63 outperforming the baseline neural network architectures commonly used for hydrological modeling and improved the performance of DLEM (Table 1). We use DLEM (PB) simulated runoff and similar preprocessing of all the input vectors for a fair assessment of the benchmark architectures.
\begin{figure}[t]
\centering
\includegraphics[width=\columnwidth]{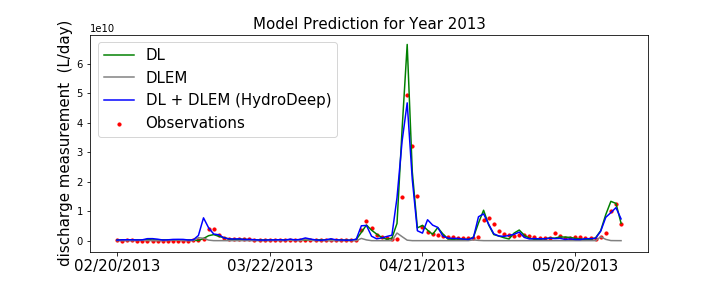} 
\caption{We present a closer look at a period from February 20, 2013 to May 20, 2013 in watershed 13 where HydroDeep, with the use of DLEM, outperforms the other models. DL achieves an NSE of 0.49 compared to HydroDeep's 0.63.}
\label{fig6}
\end{figure}
\begin{table}[b]
\centering
\resizebox{\columnwidth}{!}{
\begin{tabular}{lcrrr}
    \toprule
        \textbf{Models} & \textbf{Type} & \textbf{NSE} & \textbf{PBIAS} &  \textbf{RSR}\\
   \midrule
        HydroDeep & DL + PB & 0.63 & 10.54 & 0.61\\
        CNN & DL + PB  & 0.62 & -22.81 & 0.62\\
        LSTM & DL + PB & 0.57 & -27.21 & 0.65\\
        GRU & DL + PB & 0.50 & -20.35 & 0.71\\
        Bi-directional LSTM & DL + PB & 0.41 & -3.91 & 0.77\\
        DLEM & PB &  - 0.11 & -97.21 & 1.05\\
    \bottomrule
\end{tabular}
}
\caption{Evaluation of HydroDeep's performance compared to benchmark neural network architectures and DLEM.}
\label{table1}
\vspace{-3mm}
\end{table}
\begin{table*}[t]
\centering
\resizebox{2.05\columnwidth}{!}{
\begin{tabular}{lrrrrrrrrrrrrrrrr}
    \toprule
        \textbf{Source - Watershed 13 (grids = 29)} & {} & \multicolumn{3}{c}{\textbf{HydroDeep}} & \multicolumn{3}{c}{\textbf{T-HydroDeep-1}} & \multicolumn{3}{c}{\textbf{T-HydroDeep-2}} & \multicolumn{3}{c}{\textbf{T-HydroDeep-3}} & \multicolumn{3}{c}{\textbf{T-HydroDeep-4}}\\
        Target Watersheds/Metrics &	{No. of grids}& NSE	& PBIAS & RSR &	NSE &	PBIAS &	RSR &	NSE &	PBIAS &	RSR &	NSE &	PBIAS & RSR &	NSE &	PBIAS & RSR\\
    \midrule
        Watershed 14 (w14) & 34 & 0.27 &	-12.14 &	0.86 &	0.33 &	-27.08 &	0.82 &	0.30 &	-29.33 &	0.83 &	0.32 &	-20.03 &	0.83 &	\textbf{0.39} &	-24.68 &	0.78\\
        Watershed 15 (w15) & 39 & 0.24 &	-1.16 &	0.87 &	0.46 &	-15.76 &	0.73 &	0.47 &	-2.43 &	0.73 &	0.42 &	-8.45 &	0.76 &	\textbf{0.50} &	2.92 &	0.71\\
        Watershed 10 (w10) & 65 & 0.80 &	2.72 &	0.45 &	0.82 &	-5.08 &	0.43 &	\textbf{0.87} &	1.52 &	0.36 &	0.86 &	2.48 &	0.37 &	0.86 &	-7.44 &	0.38\\
        Watershed 4 (w4) &	61 & 0.71 &	12.37 &	0.54 &	0.76 &	-11.43 &	0.49 &	\textbf{0.82} &	-5.27 &	0.42 &	0.81 &	3.85 &	0.44 &	0.82 &	-1.14 &	0.42\\
        Watershed 23 (w23) & 32 & 0.36 &	38.05 &	0.8 &	0.45 &	9.71 &	0.74 &	0.38 &	39.56 &	0.79 &	\textbf{0.50} &	-13.38 &	0.71 &	0.46 &	40.51 &	0.73\\
    \bottomrule
\end{tabular}}
\caption{Performance comparison of untrained Hydrodeep with four Transfer Learning approaches where HydroDeep was pretrained on w13 and transferred to other watersheds of varying distance and trained (finetuned) for 20 iterations. (Supplementary Section 3.)}
\vspace{-2mm}
\label{table2}
\end{table*}
We also compare their performances to DLEM's discharge prediction. The hydrological parameters in the DLEM model are calibrated against observational river discharge flowing out of large river basins that include complex spatial heterogeneity (e.g., the Mississippi River Basin). The calibrated parameters represent an “average” condition in hydrology over a large spatial extent. However, when applied to a smaller watershed (e.g., w13 in this study), these parameters may bias the river discharge estimates due to the simple landscape. To precisely simulate river discharge at smaller watersheds, the DLEM parameters need to be individually calibrated, which is inefficient in terms of computational efforts. Our research claims that more grids in a region will increase HydroDeep's prediction power, which has been discussed more thoroughly in Section 5.4. 
\subsection{Regional Knowledge Transfer for Spatiotemporal Analysis}
Spatiotemporal variance influencing local river discharge is inevitable due to varying soil property, climate, and land usage among different regions. Training local hydrological models to capture regional spatiotemporal features require extensive training and computational resources. These regions, although they vary in local characteristics, still follow fundamental hydrological dependencies. We have experimented with four transfer learning approaches to reuse HydroDeep’s knowledge in predicting river discharge in distant watersheds. The goal was to find the best approach to transfer HydroDeep’s knowledge from one watershed to another to reduce the required training iterations on a new region and analyze the geo-spatiotemporal similarities and dissimilarities between the source and the target. In the first approach, we preserve the original HydroDeep’s spatiotemporal knowledge from the source w13 and use them solely in T-HydroDeep-1 to test new target watersheds. Secondly, we transfer the original HydroDeep’s spatiotemporal knowledge in T-HydroDeep-2 and allow it to finetune on the target. In the third and fourth approaches, we take turns in finetuning just the temporal features in T-HydroDeep-3 and the spatial features in T-HydroDeep-4. The CNN layers and the LSTM layers are responsible for learning spatial and temporal features, respectively. When we finetune one kind of the layers, we freeze the other kind to keep the originally learned features intact. Table 2 shows the observations from our experiments. The watersheds w14 and w15, although being adjacent to the source w13, are observed to have distinct spatial features as T-HydroDeep-4 proved to be the best approach for both the watersheds. We also observed that the second-best approach for w14 is T-HydroDeep-1, which means w13 and w14 have similar temporal features, but since they vary in spatial features, only the spatial layers needed to be finetuned (T-HydroDeep-4). Similarly, in w15, the result shows that the watershed has more distinct spatial features than temporal features, as T-HydroDeep-4 proved to be the best approach and T-HydroDeep-2, the second-best. However, w4 and w10, both being far from w13, show distinct spatiotemporal features as T-HydroDeep-2 shows the best performance. Note that w13 has only 29 spatial grids, whereas both w4 and w10 have 61 and 65 grids, respectively. HydroDeep’s knowledge is transferred to targets almost double in the area and achieved the best performances among all the watersheds included in our experiment. This supports our argument that more data availability will increase HydroDeep’s performance, and transfer learning approaches work in targets larger in area than the source, and can be used in analyzing spatiotemporal characteristics of watersheds. 
\section{Conclusion and Future Work}This paper illustrates a new approach to analyze unique regional geo-spatiotemporal information. DLEM's knowledge of PB mechanisms and a combination of CNN-LSTM to capture the local geo-spatiotemporal features drove HydroDeep to better performance over other baseline architectures. The extracted geospatial features steered the recurrent layers to discover the local temporal features of the watersheds. Furthermore, results show that a set of transfer learning techniques, when applied, can help analyze geo-spatiotemporal characteristics of watersheds in minimal training duration and limited computational resources.
In the future, a larger watershed with more diverse characteristics can be chosen as a source, which we believe will enable HydroDeep to work even better in transferring knowledge to new watersheds having unique features. This work can also be extended to a smaller grid-scale resolution enabling HydroDeep or similar models to capture local geo-spatiotemporal features on a more finer-scale. 

\bibliographystyle{named}
\bibliography{main}

\end{document}